\documentclass[letterpaper]{article}
\usepackage{aaai}
\usepackage{times}
\usepackage{helvet}
\usepackage{courier}
\usepackage{graphicx}
\usepackage{url}
\usepackage{booktabs}
\usepackage{amsfonts}
\usepackage{nicefrac}
\usepackage[table]{xcolor}
\usepackage{amsmath}
\begin{document}

\title{Multi-Agent Collaborative Reward Design for Enhancing Reasoning in Reinforcement Learning}
\author{Pei Yang$^{1}$, Ke Zhang$^{2}$, Ji Wang$^{3}$, Xiao Chen$^{4}$, Yuxin Tang$^{5,1}$, Eric Yang$^{1}$, Lynn AI$^{1}$, Bill Shi$^{1\dagger}$\\
$^{1}$Gradient, $^{2}$Waseda University, $^{3}$Columbia University,\\
$^{4}$Hong Kong Polytechnic University, $^{5}$Rice University\\
yangpei@gradient.network, zhangke@moegi.waseda.jp, jw4809@columbia.edu,\\
shawn.chen@connect.polyu.hk, yuxin.tang@rice.edu, ey@gradient.network,\\
lynn@gradient.network, tianyu@gradient.network
}
\maketitle

\begin{abstract}
We present Collaborative Reward Modeling (CRM), a framework that replaces a single black-box reward model with a coordinated team of specialist evaluators to improve robustness and interpretability in RLHF. Conventional reward models struggle to jointly optimize multiple, sometimes conflicting, preference dimensions (e.g., factuality, helpfulness, safety) and offer limited transparency into why a score is assigned. CRM addresses these issues by decomposing preference evaluation into domain-specific agents that each produce partial signals, alongside global evaluators such as ranker-based and embedding-similarity rewards. A centralized aggregator fuses these signals at each timestep, balancing factors like step-wise correctness, multi-agent agreement, and repetition penalties, yielding a single training reward compatible with standard RL pipelines. The policy is optimized with advantage-based updates (e.g., GAE), while a value model regresses to the aggregated reward, enabling multi-perspective reward shaping without requiring additional human annotations beyond those used to train the evaluators. We evaluate CRM on RewardBench, a benchmark suite aligned with multi-dimensional preference evaluation, demonstrating a practical, modular path to more transparent reward modeling and more stable optimization.
\end{abstract}

\noindent $\dagger$ Corresponding author

\section{Introduction}

Large language models (LLMs) are typically aligned with human preferences using reinforcement learning techniques~\cite{brown2020language,touvron2023llama,achiam2023gpt}, where a learned reward model guides the policy toward preferred behaviors. The reward model (RM) is trained on human preference data to assign scores or provide pairwise rankings for LLM outputs~\cite{stiennon2020learning,ouyang2022training}. While effective, this monolithic approach has notable limitations. First, human preferences are inherently multi-dimensional -- spanning factors such as factual accuracy, coherence, helpfulness, and safety -- yet a single scalar reward cannot easily capture trade-offs between these sometimes competing criteria. Second, conventional reward models offer limited interpretability: they provide little insight into why a given output is rated highly or poorly. This opacity makes it difficult to diagnose mistakes and increases the risk of reward hacking, where the policy learns to exploit quirks of the reward function (producing outputs that score well but are misaligned with true intent). These issues motivate a rethinking of reward modeling for safer and more transparent RLHF.

In this work, we propose a new paradigm called Collaborative Reward Modeling (CRM) that transforms the single-agent reward model into multiple specialized evaluators working in coordination. Instead of relying on one complex model to judge multiple aspects of an LLM's output, CRM employs several LLM-based reward agents, each specialized in a particular dimension of preference evaluation. For example, one agent focuses on factual correctness and reasoning validity, another on the helpfulness and coherence of the response. Each specialist evaluator reads the policy's output (including step-by-step reasoning and a final answer) and produces an interpretable partial score. In addition to these domain-specific judges, we incorporate global evaluators that provide holistic feedback -- notably a ranker-based preference score and an embedding-similarity score measuring semantic alignment with reference answers or prompts. By decomposing the reward signal into multiple components, CRM yields richer and more transparent feedback than a single scalar model. Each agent's score represents a human-understandable evaluation (e.g., factual accuracy), which collectively form a multi-dimensional portrait of the output's quality.

In particular, the proposed CRM comprises three components: (i) \textbf{Collaborative Reward Modeling}. Instead of one black-box score, distributed agents evaluate each policy rollout from complementary perspectives. Their signals include both global evaluators (e.g., ranker- or embedding-based similarity rewards) and agent-specific assessments (e.g., reasoning quality, format adherence, stability diagnostics). These are combined into a collaborative reward with adaptive weights.
(ii) \textbf{Centralized Reward Aggregation}. A centralized aggregator fuses the collaborative reward with auxiliary enhancements, balancing factors such as step-wise correctness, agreement among agents, and repetition penalties. The result is a single training reward that preserves multi-dimensional feedback while remaining compatible with standard RL pipelines.
(iii) \textbf{Policy and Value Updates}. We optimize the policy with advantage-based updates and fit a value function to the centralized reward. This design shapes learning with multi-perspective feedback, aiming to improve robustness and sample efficiency without requiring extra human annotations beyond those used to train the constituent evaluators.

We evaluate CRM on RewardBench, a benchmark suite providing datasets and tasks with multi-dimensional preference signals for evaluating end-to-end performance in settings that jointly require correctness, helpfulness, and safety. Experiments on GSM8K and RewardBench show that CRM substantially improves reasoning accuracy and robustness while maintaining or improving dialogue quality and safety.

Our contributions are as follows:

\begin{itemize}
    \item A new paradigm for reward modeling that extends RLHF with cooperative multi-agent evaluation, improving interpretability and robustness over a single black-box RM.
    \item A structured collaborative reward mechanism (MARM) with specialist evaluators and a centralized aggregator that fuses multi-dimensional, interpretable signals into a single reward usable by standard policy-gradient methods.
    \item Significant gains on complex reasoning tasks, with higher accuracy and stability than single-RM baselines while preserving fluency and safety, demonstrating the effectiveness of multi-perspective reward shaping.
\end{itemize}

\section{Related Work}
\label{sec:related}

\paragraph{Reward Modeling and RLHF.} Reinforcement learning from human feedback (RLHF) has become the dominant paradigm for aligning large language models (LLMs) with human preferences~\cite{Christiano2017,Stiennon2020,Ouyang2022,Bai2022}. In this pipeline, a scalar reward model is trained on human preference data to guide policy updates via RL. While successful in applications like InstructGPT~\cite{Ouyang2022} and GPT-4~\cite{OpenAI2023}, this scalar setup offers limited transparency, making the system vulnerable to reward hacking~\cite{Skalse2022} and misalignment under distribution shift~\cite{Coste2023}. To mitigate this, recent work explores ensemble-based reward models~\cite{Coste2023} and multi-objective formulations~\cite{Wang2024}, which decompose feedback into interpretable dimensions (e.g., helpfulness, honesty, verbosity) and re-aggregate them via learned mixtures. Another approach is to make reward models self-reflective, as in Critique-out-Loud~\cite{Ankner2024}, which outputs both scores and critiques to improve interpretability.

\paragraph{Multi-Agent and Structured Evaluation.} Structured evaluation and multi-agent feedback have emerged as promising alternatives to monolithic reward modeling. AI Safety via Debate~\cite{Irving2018} pioneered the idea of letting two models argue and a third evaluate, introducing competition as a feedback mechanism. Multi-role or multi-agent feedback has since been applied in RLAIF-style setups~\cite{Cheng2024}, where agents simulate reviewers or judges from diverse perspectives. ChatEval~\cite{Chan2023} aggregates multiple LLMs into a referee panel that debates and votes on completions, improving human alignment. CRM differs from these by using agents not only at evaluation time but also in reward modeling, integrating specialist agents (e.g., for reasoning correctness, formatting, or statistical diversity) as real-time contributors to the reward signal. This allows structure-aware, multi-view feedback during training, complementing recent techniques like GRPO~\cite{Zheng2024}, SPIN~\cite{Sun2023SPIN}, and RAFT~\cite{Yang2023RAFT} that focus on fine-grained or step-wise feedback signals.

\begin{figure*}[t]
    \centering
    \includegraphics[width=0.85\textwidth]{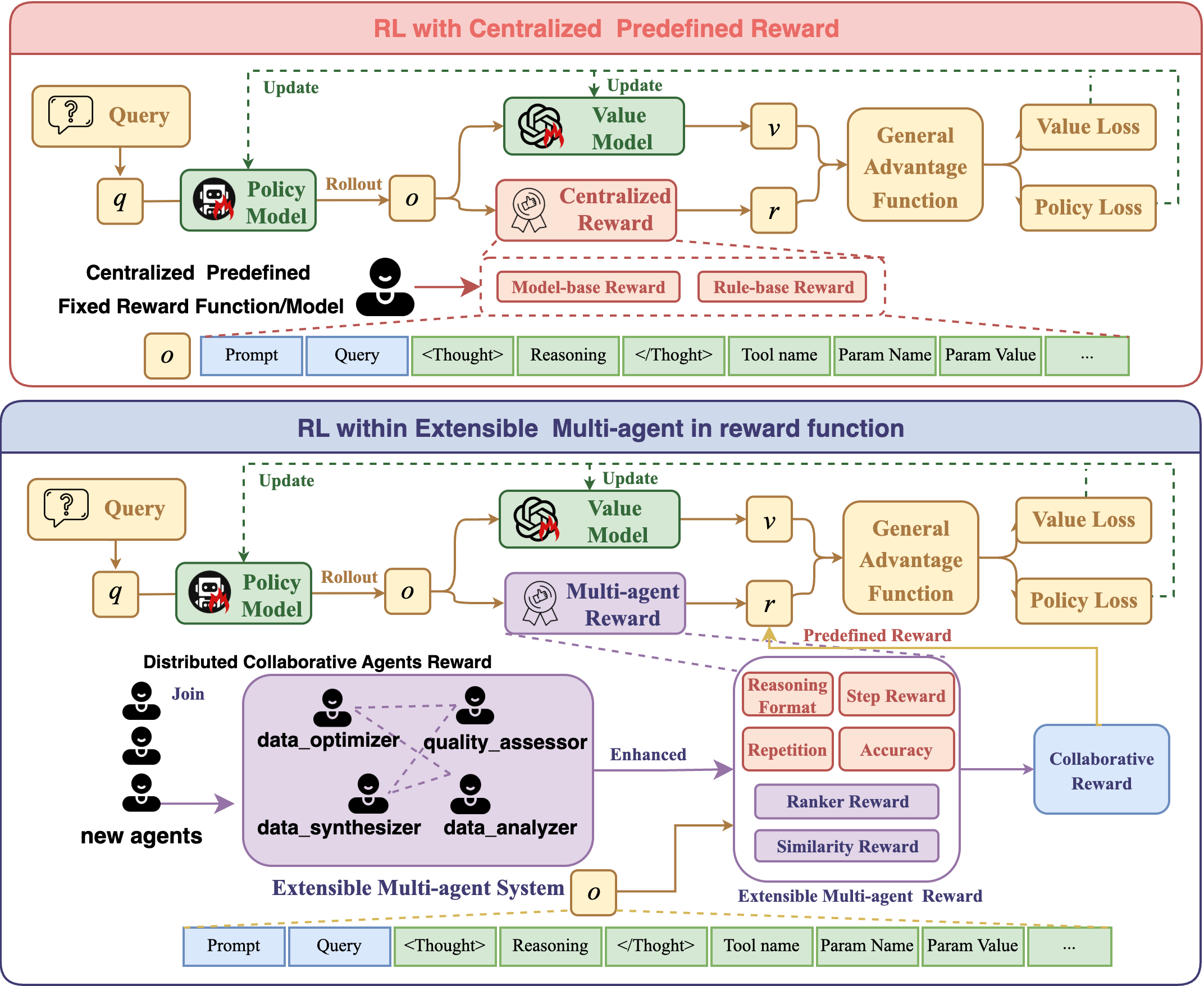}
    \caption{Architecture of CRM. In comparison with the predefined and fixed reward function in the conventional method, CRM leverages a multi-agent system to build an extensible intelligent reward function.}
    \label{fig:Method}
\end{figure*}

\section{Methodology}
\label{sec:method}

In this section, we present our \textit{Collaborative Reward Model via Reinforcement Learning}, which integrates distributed agents into the reward function to enhance policy optimization. Unlike conventional reinforcement learning (RL) pipelines that rely on a monolithic scalar reward, our framework leverages a collaborative reward mechanism, enabling fine-grained evaluation and adaptive feedback. The overall workflow is illustrated in Fig.~\ref{fig:Method}.

\subsection{Problem Formulation}

We consider a learning paradigm where a large policy model 
$\pi_\theta$ generates structured rollouts $o = \pi_\theta(x)$ 
in response to prompts $x$. 
Each rollout may contain multi-step reasoning traces and final answers. 
Rather than optimizing a fixed scalar reward, our goal is to learn from 
a \emph{multi-dimensional evaluation space} provided by a group of collaborative agents.

Formally, each rollout $o$ is evaluated by a set of agents 
$\mathcal{A} = \{a_1, a_2, \dots, a_K\}$, 
where agent $a_i$ emits a score $R_i(o)$ for a specific evaluation dimension 
(e.g., accuracy, coherence, diversity, stability).
Additionally, two global evaluators, 
$R_{\mathrm{ranker}}(o)$ and $R_{\mathrm{similarity}}(o)$, 
capture overall preference alignment and semantic consistency with reference outputs.

The objective of training is to optimize $\pi_\theta$ such that
its expected aggregated reward is maximized:
\begin{equation}
\max_{\theta}\; 
\mathbb{E}_{x\sim \mathcal{D}} 
\Big[
F\!\big(
\alpha R_{\mathrm{ranker}}(o) 
+ \beta R_{\mathrm{similarity}}(o)
+ \sum_{i=1}^{K} \lambda_i R_i(o)
\big)
\Big],
\end{equation}
where $F(\cdot)$ denotes the central aggregator transforming 
heterogeneous signals into a scalar reward, 
and $\{\alpha, \beta, \lambda_i\}$ are adaptive weights learned or tuned during training.  
This formulation generalizes traditional RLHF by replacing the single black-box 
reward model with a structured, interpretable, and extensible multi-agent evaluation process.

\subsection{Collaborative Reward Modeling}

We propose a \textbf{Collaborative Reward Model (CRM)} that restructures post-training into a distributed, feedback-driven optimization process. Rather than relying on a single black-box reward model, CRM introduces a team of \textbf{specialist agents} that cooperatively evaluate large-model rollouts from complementary perspectives, yielding multi-dimensional feedback later fused into a scalar reward compatible with standard GRPO or PPO optimization (see Figure~\ref{fig:Method}). Specifically, the \textbf{Data Optimizer} quantifies rollout efficiency and diversity, penalizing redundant reasoning traces while encouraging exploration balance; the \textbf{Quality Assessor} provides fine-grained judgments on reasoning accuracy, factual consistency, and logical coherence across intermediate steps; the \textbf{Data Synthesizer} augments supervision by injecting synthetic perturbations and integrating external knowledge, thereby improving robustness and domain generalization; and the \textbf{Data Analyzer} continuously monitors statistical trends of the reward signals, enforcing stability and preventing collapse or mode drift.

\subsection{Reward Function Design}

\begin{figure}[htbp]
    \centering
    \includegraphics[width=0.95\linewidth]{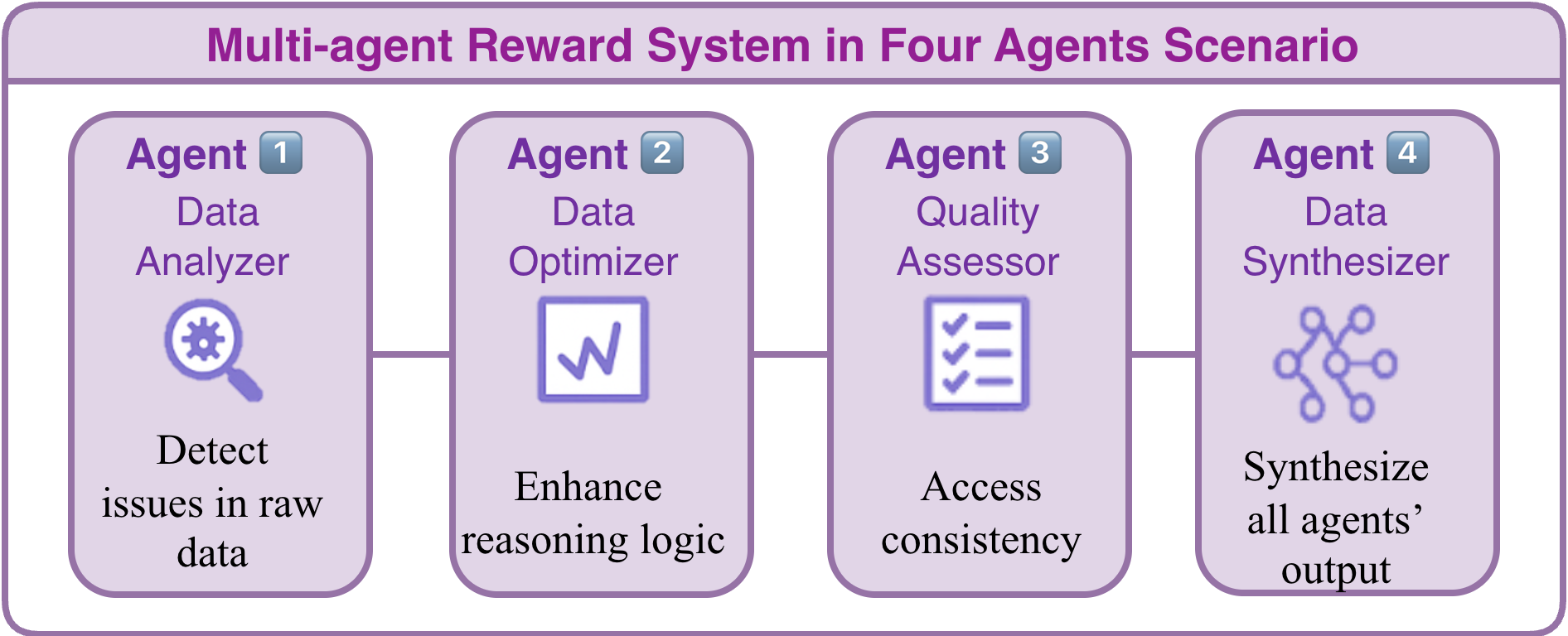}
    \caption{Decomposition of collaborative reward roles.}
    \label{fig:placeholder}
\end{figure}

A key instantiation of the CRM framework is the Multi-Agent Reward Model (MARM), which operationalizes the concept through multi-dimensional feedback (Figure~\ref{fig:Method}). 
Each agent contributes interpretable reward components that together form a unified, differentiable objective for post-training. 
The full reward integrates both step-level and model-level signals as follows.

\paragraph{Step-level Rewards.}
To capture reasoning progress, we assign token-level signals for intermediate steps. 
The \textbf{Outcome Reward} verifies whether partial reasoning aligns with intermediate expectations, 
while the \textbf{Enhanced Data Reward} leverages augmented or counterfactual samples generated by the Data Synthesizer to provide stronger supervision. 
These components guide the model to maintain logical validity throughout multi-step derivations rather than optimizing only the final output.

\paragraph{Model-based Rewards.}
At the sequence level, we employ semantic similarity as a model-aware signal. 
Following common practice in reward modeling, we use the \texttt{all-MiniLM-L6-v2} encoder to compute cosine similarity between the predicted and reference embeddings:
\begin{equation}
R_{\text{sim}} = \cos(\mathbf{h}_{\text{pred}}, \mathbf{h}_{\text{ref}}),
\end{equation}
where $\mathbf{h}_{\text{pred}}$ and $\mathbf{h}_{\text{ref}}$ denote the sentence-level representations of the generated and reference responses. 
This signal captures semantic closeness even when surface forms differ, complementing the more discrete correctness-based rewards.

\paragraph{Multi-dimensional Evaluation.}
Beyond step- and model-level scores, additional evaluators encode structural and behavioral preferences:
\textbf{Accuracy Reward} ($R_{\text{acc}}$) validates mathematical equivalence through symbolic comparison (\texttt{latex2sympy2}, \texttt{math\_verify}); 
\textbf{Format Reward} ($R_{\text{fmt}}$) enforces adherence to the reasoning format defined by \texttt{<think>} and \texttt{<answer>} tags; 
\textbf{Reasoning-Step Reward} ($R_{\text{step}}$) encourages organized, interpretable multi-step explanations; 
\textbf{Cosine-Scaled Reward} ($R_{\text{cs}}$) modulates accuracy reward by completion length to prevent verbosity; 
and \textbf{Repetition Penalty} ($R_{\text{rep}}$) penalizes $n$-gram redundancy and degenerate loops detected by the Data Analyzer.

All signals are unified through the collaborative weighting mechanism:
\begin{equation}
R_{\text{collab}} = 
\alpha R_{\text{acc}} 
+ \beta R_{\text{sim}} 
+ \gamma R_{\text{fmt}} 
+ \delta R_{\text{step}} 
- \eta R_{\text{rep}},
\end{equation}
where the coefficients $(\alpha, \beta, \gamma, \delta, \eta)$ are tuned empirically to balance factual correctness, reasoning clarity, and linguistic fluency. 
This formulation ensures that the collaborative reward remains interpretable, extensible, and fully compatible with standard policy optimization objectives.

\subsection{Reward Aggregation and Policy Updates}

To integrate these signals, we adopt a centralized aggregation mechanism:
\begin{equation}
r_t = \mathcal{F}\Big(R_{\text{collab}}(o_t), R_{\text{enhanced}}(o_t)\Big),
\end{equation}
where $o_t$ denotes the policy rollout at timestep $t$, and $\mathcal{F}$ is a non-linear fusion operator that balances reasoning format, accuracy, and repetition penalties. The enhanced rewards include auxiliary dimensions such as step-wise correctness and multi-agent agreement.

During training, the policy model $\pi_\theta$ is updated using generalized advantage estimation (GAE), while the value model $V_\phi$ is optimized via regression on centralized rewards:
\begin{align}
\mathcal{L}_{\text{policy}} &= - \mathbb{E}_t \big[ \hat{A}_t \log \pi_\theta(a_t|s_t) \big], \\
\mathcal{L}_{\text{value}} &= \mathbb{E}_t \big[ (V_\phi(s_t) - r_t)^2 \big],
\end{align}
where $\hat{A}_t$ is the advantage function. The collaborative design ensures that policy learning is guided by multi-perspective reward shaping, which improves robustness and sample efficiency.

\subsection{Discussion}

The CRM redefines reward modeling as an adaptive, interpretable, and extensible ecosystem rather than a fixed scoring oracle. By coordinating distributed evaluators through a centralized fusion mechanism, it achieves transparency and robustness in large-model optimization. The modular design allows new evaluators to be introduced as plug-in agents, providing a scalable pathway toward self-regularizing and interpretable reward alignment.

\section{Experiments}
\label{sec:experiments}

\begin{table*}[t]
\centering
\caption{Result of MARM in RewardBench, Math and GSM8K}
\label{tab:result}
\begin{tabular}{lcccccc}
\toprule
 Methods & Chat & Chat Hard & Safety & Reasoning & Math & GSM8K \\
\midrule
\rowcolor{gray!30} \multicolumn{7}{c}{\textit{Two Agents~(Data Analyzer + Data Optimizer)}} \\
\midrule
Qwen2.5-0.5B-ins & 0.193 & \textbf{0.561} & 0.561 & 0.598 & 0.139 & 0.08\% \\
Qwen2.5-0.5B (GRPO) & 0.190 & 0.557 & 0.553 & \textbf{0.659} & \textbf{0.149} & 19.64\% \\
MARM & 0.182 & 0.545 & \textbf{0.566} & 0.423 & 0.136 & 22.16\% \\
MARM(emb) & \textbf{0.198} & \textbf{0.561} & 0.536 & 0.567 & 0.131 & \textbf{22.33\%} \\
\midrule
\rowcolor{gray!30}\multicolumn{7}{c}{\textit{Three Agents ~(Data Analyzer + Data Optimizer + Quality Assessor)}} \\
\midrule
Qwen2.5-0.5B-ins & 0.193 & 0.561 & 0.561 & 0.598 & 0.139 & 0.08\% \\
Qwen2.5-0.5B (GRPO) & 0.190 & 0.557 & 0.553 & \textbf{0.659} & \textbf{0.149} & 19.64\% \\
MARM & 0.190 & \textbf{0.567} & 0.538 & 0.398 & 0.143 & 22.87\% \\
MARM(emb) & \textbf{0.199} & 0.532 & \textbf{0.570} & 0.637 & 0.141 & \textbf{23.15\%} \\
\midrule
\rowcolor{gray!30}\multicolumn{7}{c}{\textit{Four Agents~(Data Analyzer + Data Optimizer + Quality Assessor + Data Synthesizer)}} \\
\midrule
Qwen2.5-0.5B-ins & \textbf{0.193} & 0.561 & 0.561 & 0.598 & 0.139 & 0.08\% \\
Qwen2.5-0.5B (GRPO) & 0.190 & 0.557 & 0.553 & \textbf{0.659} & 0.149 & 19.64\% \\
MARM & 0.182 & \textbf{0.568} & 0.527 & 0.610 & \textbf{0.192} & \textbf{29.87\%} \\
MARM(emb) & 0.179 & 0.557 & \textbf{0.573} & 0.578 & 0.152 & 27.60\% \\
\bottomrule
\end{tabular}
\end{table*}

\subsection{Experimental Setup}
\label{sec:model_data}

This study evaluates the proposed collaborative reward modeling (CRM) framework using the RewardBench benchmark and two common datasets---GSM8K and Math for mathematical reasoning. The base model is \textbf{Qwen/Qwen2.5-0.5B-Instruct}, containing approximately 494M parameters. Training utilizes the \textbf{AI-MO/NuminaMath-TIR} dataset, which includes 3,800 samples in the training set and 99 samples in the test set. All experiments are conducted under the Generalized Reinforcement Policy Optimization (GRPO) framework, which supports the integration of multiple reward sources and ensures stable policy updates during training.

We compare collaborative configurations involving two, three, and four agents:
\begin{itemize}
    \item \textbf{Two Agents}: Data Analyzer + Data Optimizer
    \item \textbf{Three Agents}: Data Analyzer + Data Optimizer + Quality Assessor
    \item \textbf{Four Agents}: Data Analyzer + Data Optimizer + Quality Assessor + Data Synthesizer
\end{itemize}

Each agent contributes distinct evaluation signals such as logical consistency, factuality, step-wise reasoning accuracy, and stylistic robustness. To standardize reasoning and output, we employ a \textbf{system prompt}, which enforces explicit reasoning traces within \texttt{<think>...</think>} tags and final answers within \texttt{<answer>...</answer>} tags. This structured prompting allows the reward model to evaluate both reasoning quality and final correctness.

\subsection{Results and Analysis}

The experimental results presented in Table~\ref{tab:result} demonstrate the efficiency of collaborative reward modeling under different agent configurations.
First, in the process from two-agent to three-agent collaboration yields improvements across reasoning-related metrics, particularly in the \textit{Reasoning} and \textit{Math} categories. The addition of the \textbf{Quality Assessor} introduces more fine-grained evaluative feedback, allowing the reward function to better capture structural coherence and step-wise logical soundness. This leads to a notable increase in reasoning accuracy, with the MARM(emb) variant improving from 0.567 to 0.637, and a similar upward trend observed in the GSM8K benchmark (from 22.33\% to 23.15\%). These results suggest that multi-perspective evaluation promotes better alignment between reasoning structure and correctness.

Second, when expanding to the four-agent configuration by introducing the \textbf{Data Synthesizer}, further gains are observed, especially on tasks requiring compositional reasoning and factual synthesis. The MARM variant achieves the best overall performance, reaching 0.610 on the reasoning dimension and 29.87\% accuracy on GSM8K. This indicates that the synthesizer enhances model generalization by mitigating local overfitting and improving the semantic completeness of intermediate reasoning chains. Moreover, the MARM(emb) variant maintains relatively stable results across multiple metrics, demonstrating robustness in embedding-level reward aggregation and its potential to stabilize optimization dynamics.

Third, while the improvements on general dialogue tasks such as \textit{Chat} and \textit{Chat Hard} are comparatively moderate, the overall consistency of performance across these categories indicates that the collaborative framework does not compromise fluency or safety. The reward fusion mechanism effectively balances multi-dimensional objectives---ensuring that reasoning-oriented optimization does not lead to degradation in conversational quality or stylistic alignment.

Finally, the performance differences among Qwen2.5-0.5B (GRPO), MARM, and MARM(emb) highlight the importance of \textbf{reward aggregation strategies}. The reranking-based approach consistently outperforms other variants, suggesting that explicit preference modeling and pairwise ranking contribute to more discriminative reward shaping. Embedding-based fusion, while slightly less performant in high-precision reasoning, demonstrates better stability and scalability for complex multi-agent coordination.

Overall, these findings confirm that collaborative reward modeling provides a structured and extensible mechanism for integrating heterogeneous evaluators. By aligning multi-agent feedback through a centralized aggregation operator, the framework achieves significant gains in reasoning robustness and mathematical precision without sacrificing general conversational quality.

\section{Conclusion}

In this work, we introduced the Collaborative Reward Modeling (CRM) framework, which redefines reward modeling as a multi-agent evaluation process rather than a single black-box oracle. We instantiated this framework as MARM (Multi-Agent Reward Model), decomposing preference assessment into specialized evaluators and integrating their signals through a centralized aggregation mechanism.

Comprehensive experiments on RewardBench, Math, and GSM8K demonstrate that multi-agent collaboration substantially enhances reasoning accuracy, mathematical precision, and overall stability without compromising conversational quality. The introduction of roles such as the Quality Assessor and Data Synthesizer further improves consistency and generalization, highlighting the advantage of domain-specific decomposition and coordinated feedback in reward modeling.

Beyond performance gains, CRM offers a scalable and modular design that supports the integration of new evaluators as plug-in agents, making it compatible with existing RLHF pipelines. This work lays the groundwork for the next generation of interpretable and self-regularizing reward systems, bridging the gap between human-aligned evaluation and efficient large-model optimization.

\bibliographystyle{aaai}
\bibliography{ref}

\end{document}